\documentclass[pdflatex,sn-mathphys-num]{sn-jnl}% Math and Physical Sciences Numbered Reference Style
\usepackage{graphicx}%
\usepackage{multirow}%
\usepackage{amsmath,amssymb,amsfonts}%
\usepackage{amsthm}%
\usepackage{mathrsfs}%
\usepackage[title]{appendix}%
\usepackage{xcolor}%
\usepackage{textcomp}%
\usepackage{manyfoot}%
\usepackage{booktabs}%
\usepackage{algorithm}%
\usepackage{algorithmicx}%
\usepackage{algpseudocode}%
\usepackage{listings}%
\usepackage{booktabs}
\usepackage{xcolor}
\usepackage{multirow}

\usepackage{makecell}
\usepackage{pifont}
\newcommand{\cmark}{\ding{51}} % tick ✔
\usepackage{subcaption}

\theoremstyle{thmstyleone}%
\theoremstyle{thmstyletwo}%

\theoremstyle{thmstylethree}%

\begin{document}

\title[Article Title]{Head-Pose-Aware Visual Speech Recognition with FiLM Modulation}
% Seeing Speech from Every Angle: Head-Pose-Aware Visual Speech Recognition with FiLM Modulation


\author*[1]{\fnm{Matthew Kit Khinn} \sur{Teng}}\email{teng.khinn-matthew178@mail.kyutech.jp}

\author[1]{\fnm{Haibo} \sur{Zhang}}\email{haiboz@ai.kyutech.ac.jp}

\author[1]{\fnm{Takeshi} \sur{Saitoh}}\email{saitoh@ai.kyutech.ac.jp}

\affil*[1]{\orgdiv{Department of Artificial Intelligence}, \orgname{Kyushu Institute of Technology}, \orgaddress{ \city{Iizuka}, \postcode{820-8502}, \state{Fukuoka}, \country{Japan}}}

\abstract{Visual Speech Recognition (VSR) aims to recognize speech from visual cues such as lip movements. Still, its performance is fundamentally limited by viseme ambiguity and pose-induced variations that introduce geometric distortions and occlusions. Existing approaches mainly rely on linguistic context or implicit invariance, leaving visual representations insufficiently robust under non-frontal views. In this work, we propose a pose-aware phoneme-level framework, termed HP-VSR-ResFiLM, that explicitly incorporates head-pose information into visual feature extraction. The proposed framework adopts a two-stage pipeline consisting of a pose-conditioned visual encoder in Stage~1 and a pretrained NLLB language model in Stage~2 for phoneme-to-text reconstruction. Specifically, Stage~1 incorporates a pose-conditioned residual Feature-wise Linear Modulation (FiLM) block after the 2D CNN frontend to refine visual representations adaptively using head-pose information. Experiments on LRS2 and LRS3 demonstrate that HP-VSR-ResFiLM achieves competitive performance under comparable training conditions, attaining word error rates (WER) of 24.7\% and 30.3\%, respectively, without relying on additional training data. Comprehensive ablation studies reveal that different FiLM modulation strategies exhibit complementary strengths: HP-VSR-FiLMFuse (L4) achieves the best overall performance on LRS2, whereas the proposed HP-VSR-ResFiLM provides the greatest improvements under large head-pose variations on LRS3 and consistently outperforms previous pose-aware methods for high-yaw samples. These results demonstrate the effectiveness of explicit pose-conditioned feature modulation for robust visual speech recognition in unconstrained settings.}


\maketitle

\section{Introduction}\label{intro}

Human speech communication is inherently multimodal, where visual cues from lip movements and facial dynamics complement acoustic information \cite{sumby_1954}. In challenging acoustic environments, such as noisy public spaces or far-field recording conditions, visual signals become particularly important for reliable recognition. Motivated by this observation, VSR, also known as lip-reading, aims to recognize spoken content solely from visual input. Recent advances in deep learning have significantly improved VSR performance, enabling end-to-end learning from large-scale in-the-wild datasets such as LRS2 and LRS3. These developments have positioned VSR as a promising alternative or complement to traditional audio-based automatic speech recognition systems. Modern VSR architectures have evolved from early spatio-temporal convolutional and recurrent models to Transformer-based frameworks \cite{ma_2023avsr,ma_2022multilanguage} with sequence-to-sequence decoding. Large-scale datasets have facilitated the adoption of deeper visual backbones, self-supervised pretraining, and language-model-based decoding strategies.

\subsection{Problem and Motivation}

Despite recent progress in visual speech recognition, several fundamental challenges remain unresolved. In particular, existing approaches struggle to handle inherent visual ambiguities, underutilize pose-related information, and lack effective mechanisms for incorporating such cues into visual feature learning. To address these limitations, we identify the following key problems that this work aims to solve:

\begin{enumerate}

    \item \textbf{Viseme Ambiguity and the Limits of Context-Only Disambiguation.}
    A fundamental bottleneck in visual speech recognition is viseme ambiguity, where acoustically distinct phonemes — such as /b/ vs. /p/ in "back" vs. "pack" — produce visually indistinguishable lip configurations. Existing approaches address this through phoneme-centric modeling and large language model integration, both of which exploit linguistic context to infer the intended utterance. While effective in high-context settings, these strategies are unreliable when ambiguous phonemes occur in low-context or semantically balanced environments, where linguistic priors offer insufficient discriminative signal.
    
    Critically, neither strategy resolves ambiguity at the level of visual features — the representations fed into downstream decoders remain equally ambiguous regardless of the language model's capacity. We argue that this limitation can be partially mitigated during feature extraction by incorporating complementary geometric information. Head pose introduces structured geometric variations in lip appearance, revealing cues such as lateral contours and oral cavity depth that are absent in frontal views. This suggests that resolving ambiguity requires improving the quality of visual representations themselves, motivating the use of complementary geometric cues.
    
    \item \textbf{Head-Pose Variation as an Underexploited Modeling Factor.}
    Although large-scale in-the-wild benchmarks such as LRS2 and LRS3 naturally contain substantial head-pose variation, this variability is rarely treated as an explicit modeling factor. Existing approaches typically address pose variation in two ways: either by normalizing faces toward a canonical frontal view during preprocessing, or by relying on large-scale training data to learn pose-invariant representations implicitly. The former reduces pose-dependent geometric variability by treating it as incidental noise, while the latter assumes that sufficiently large models and datasets can capture pose effects without explicit supervision.
    
    However, pose-induced changes are not random; they introduce structured variations in lip appearance, articulation visibility, and oral cavity geometry that directly influence the discriminability of visual speech features. As a result, models may learn compressed representations that fail to capture pose-dependent cues.
    
    Our empirical analysis of head-pose distributions in LRS2 and LRS3 (see Section~\ref{headpose_statistic}) shows that non-frontal poses are prevalent, indicating that pose variation is not an edge case but a fundamental characteristic of real-world VSR data. This observation motivates moving beyond implicit handling and treating head pose as an explicit, informative signal within the modeling process rather than a preprocessing artifact.
    
    \item \textbf{Limitations of implicit and additive pose integration.}
    Existing methods that address pose variation rely on either multi-view modeling under controlled viewpoint supervision or augmentation-based strategies that simulate pose variability through synthetic transformations. While these approaches improve robustness within their respective settings, neither provides a general solution for unconstrained in-the-wild conditions. Multi-view designs depend on discrete, predefined viewpoints and do not extend to continuously varying poses within a single video stream. Augmentation strategies treat pose as a nuisance factor to be neutralized, discarding the discriminative geometric cues that non-frontal views carry rather than modeling them explicitly.
    
    A deeper representational issue compounds this limitation: head pose does not affect visual features uniformly. Pose-induced distortions are spatially localized and channel-specific, manifesting differently across network depths — altering lip boundary sharpness, oral aperture structure, and lateral contour visibility in ways that depend jointly on viewing angle and feature abstraction level. Strategies that handle pose globally or implicitly are insufficient to capture these layer-dependent and spatially varying effects.
    
    This motivates conditioning the visual encoder on head pose through a mechanism capable of depth-adaptive and channel-selective feature refinement. We adopt FiLM-based affine modulation, which learns to rescale and shift intermediate feature map channels as a function of instantaneous pose, providing a principled and flexible mechanism for pose-aware visual feature adaptation that neither augmentation nor multi-view alignment can offer. These observations suggest that head pose should be treated not as a nuisance factor, but as a fundamental signal for visual speech representation learning.

\end{enumerate}

\subsection{Our Contribution}
In this paper, we propose a pose-aware modulation framework, HP-VSR-ResFiLM, that incorporates a single pose-conditioned residual FiLM block after the 2D CNN frontend, enabling adaptive refinement of visual representations under varying head-pose conditions. Extensive experiments on large-scale benchmarks demonstrate that explicit pose-aware modulation significantly improves performance under unconstrained head movements. Our approach achieves substantial reductions in WER on both LRS2 and LRS3, with particularly pronounced gains on the more diverse LRS3 dataset. These results highlight the importance of explicitly modeling geometric variability in VSR and suggest that conditional feature modulation provides a principled mechanism for improving visual speech recognition.

In this work, we make the following contributions:

\begin{itemize}

\item \textbf{Reframing head pose as a fundamental signal in VSR.}
We show that head pose introduces structured geometric variations that directly affect visual speech representations, challenging the common assumption that pose should be normalized or treated as noise.

\item \textbf{Pose-conditioned residual feature modulation for VSR.}
We propose HP-VSR-ResFiLM, which incorporates a single pose-conditioned residual FiLM block after the visual frontend to refine visual representations under varying head-pose conditions adaptively. %, improving robustness with minimal architectural modification.

\item \textbf{Comprehensive analysis of pose-aware modulation strategies.}
Through controlled experiments and pose-stratified evaluations, we demonstrate that the proposed single residual FiLM block achieves the best overall WER performance across mixed-pose scenarios. Furthermore, deeper modulation at ResNet-18 Layers~3 and~4 yields larger improvements for samples with yaw angles greater than $30^\circ$, while exhibiting comparatively lower effectiveness under near-frontal pose conditions. These findings highlight the trade-off between pose-specific robustness and general pose generalization across different modulation depths.

\end{itemize}

% In this work, we make the following contributions:

% \begin{itemize}

%     \item \textbf{A two-stage phoneme-based VSR framework for viseme disambiguation.}
%     We propose a two-stage formulation that decouples pose-aware phoneme prediction from language model-based sentence reconstruction. This design improves visual-level discriminability through pose-conditioned feature extraction before applying linguistic inference, enabling more reliable disambiguation in low-context and semantically ambiguous settings.

%     \item \textbf{Modeling and empirical analysis of head-pose dynamics.}
%     We conduct a systematic analysis of head-pose distributions in LRS2 and LRS3, showing that non-frontal poses are prevalent and that standard preprocessing suppresses informative geometric cues. To capture this variation, we introduce a dedicated temporal head-pose encoder and perform evaluations across pose ranges, establishing head pose as a meaningful factor in VSR.
    
%     \item \textbf{Pose-conditioned visual feature modulation via FiLM.}
%     We introduce HP-VSR-FiLM, which integrates head-pose signals into the visual backbone via FiLM-based modulation. This enables adaptive, layer-wise refinement of intermediate representations and provides an effective mechanism for handling pose-induced distortions in unconstrained settings.
    
% \end{itemize}

\section{Related Works}\label{related}

\subsection{Visual Speech Recognition in the Wild}

Deep learning has substantially advanced VSR, evolving from early spatio-temporal convolutional models to attention-based and Transformer architectures. Initial end-to-end approaches combined convolutional encoders with recurrent networks to model temporal lip dynamics \cite{stafylakis_2017,petridis_2017_lstm,wand_2016lstm,assael_2016lipnet}. Subsequently, the availability of large-scale datasets such as GRID \cite{cooke_2006grid}, LRW \cite{chung_2017lrw}, LRS2 \cite{son_2017lrs2}, and LRS3 \cite{afouras_2018lrs3} enabled the development of more powerful architectures leveraging deeper convolutional backbones and sequence-to-sequence modeling. More recent works employ Transformer-based encoders and self-supervised pretraining to capture long-range dependencies and improve generalization in unconstrained settings.

Beyond purely visual modeling, several approaches incorporate audio information to guide visual representation learning. These methods range from direct audio-visual fusion \cite{serdyuk_2022,hu_acl_2023,hu_2023avsr} to synthetic pretraining with audio supervision \cite{ma_2023avsr,liu_2023synthvsr}. In this context, AlignVSR \cite{liu_2025alignvsr} introduces a cross-modal alignment framework that exploits both global and local audio-visual correspondence to improve visual-to-text inference, demonstrating the effectiveness of auxiliary audio supervision. While such approaches are effective, they rely on audio signals or pre-trained ASR models, which may not be available in scenarios where lip-reading is most critical.

In contrast, VSR focuses exclusively on visual input during training \cite{zhang_cvpr_2019}. Recent work further explores adapting visual encoders for pre-trained audio-based VSR models by aligning lip features with audio latent spaces \cite{djilali_2023lip2vec,laux_2024litevsr}. Although this approach benefits from strong linguistic priors learned during ASR pretraining, it typically requires substantial data to learn reliable visual representations that can bridge modality gaps.

To improve sentence-level modeling, phoneme-centric strategies introduce intermediate linguistic representations. Prajwal et al.~\cite{prajwal_2022subword} propose a Transformer-based encoder–decoder architecture that captures sub-word units through visual attention mechanisms. Earlier phoneme-based approaches \cite{elBialy_CAAI_2023} explicitly model phonetic units before word-level decoding, aiming to enhance interpretability and linguistic structure. More recently, LLMs have been incorporated into lip-reading pipelines. Yeo et al.~\cite{yeo_2024,yeo_aaai_2025} directly project visual features into LLM embedding spaces, leveraging powerful language priors for sentence reconstruction. Similarly, Thomas et al.~\cite{thomas_2025vallr} introduce VALLR, a two-stage framework that first predicts phonemes from visual input and subsequently reconstructs coherent sentences using a decoder-only LLM.

While these LLM-based approaches benefit from strong language modeling capabilities, decoder-only architectures rely heavily on the quality of intermediate phoneme predictions, and their unidirectional decoding structure may limit refinement of noisy representations under challenging visual conditions. Encoder–decoder LLM architectures, by contrast, provide context-aware encoding mechanisms that can mitigate inconsistencies in intermediate predictions through bidirectional contextual modeling \cite{eickhoff_springer_2023}. This architectural distinction suggests potential performance advantages when integrating phoneme-level representations into sentence reconstruction frameworks.

In parallel, landmark-based approaches explore geometric modeling of lip motion. Graph convolutional networks have demonstrated effectiveness in modeling non-Euclidean structures such as facial landmarks. Liu et al.~\cite{liu_2020stgcn} introduced ST-GCN for lip-reading, leveraging a fixed graph adjacency matrix to capture spatial-temporal dynamics of lip landmarks. Building upon this framework, Sheng et al.~\cite{sheng_2021asstgcn} proposed ASST-GCN, which incorporates adaptive graph learning and evaluates performance on large-scale datasets, including LRS2 and LRS3. These works highlight the complementary role of geometric representations alongside appearance-based features.

Collectively, modern VSR research has progressed from spatio-temporal convolutional modeling to Transformer-based architectures, multimodal supervision, phoneme-centric decoding, LLM integration, and graph-based geometric reasoning. These advances have substantially improved recognition accuracy on large-scale in-the-wild benchmarks such as LRS2 and LRS3. However, despite these improvements, geometric variations such as head pose are typically not modeled explicitly in current VSR systems.

\noindent\textbf{Underexplored Role of Head-Pose Variation in VSR.}
Despite the diversity of in-the-wild benchmarks, head-pose variation remains insufficiently characterized in current VSR systems. Large-scale datasets such as LRS2 and LRS3 exhibit substantial real-world variability, including illumination changes, motion blur, occlusions, and head movements. However, most VSR pipelines rely on face detection and alignment procedures that normalize faces toward a canonical frontal configuration~\cite{afouras_2018lrs3}. While effective for reducing appearance variation, such preprocessing implicitly suppresses pose diversity, limiting the model’s ability to capture pose-dependent articulation patterns. As a result, the impact of head-pose variation on large-scale VSR performance remains insufficiently characterized.

\subsection{Pose-Aware Visual Speech Recognition}

Head pose variation introduces structured geometric transformations that can significantly alter the spatial configuration of lip regions and, consequently, affect visual speech representations. While pose variability is naturally present in large-scale in-the-wild datasets, it has been more explicitly investigated in controlled multi-view settings.

Datasets such as OuluVS2 \cite{anina_2015ouluvs2} provide synchronized recordings from multiple viewing angles, enabling the development of view-specific encoders \cite{ma_IEEE_2021,isobe_2021mdpi} and pose-adaptive universal models \cite{maeda_2021apsipa}. These approaches either learn separate representations for distinct viewpoints or construct shared latent spaces that align multi-view observations. Such designs demonstrate that explicitly modeling viewpoint differences can improve recognition performance under predefined angular variations.

In addition to multi-view learning, several studies have attempted to simulate pose variability through data augmentation. These include 3D morphable model-based pose synthesis \cite{cheng_2020icassp}, synthetic data generation \cite{hao_2025lipgen}, and visual perturbation techniques \cite{fernandez_2023sparsevsr}. While these methods improve performance to specific distortions, they often depend on synthetic transformations that may introduce distribution shifts relative to real-world data. Moreover, augmentation-based strategies typically address pose variation implicitly rather than modeling it as a structured factor in the representation-learning process.

\noindent\textbf{Limitations of Implicit Head-Pose Integration.}
Prior work has explored head-pose estimation and its integration into visual perception tasks. However, explicit modeling of head pose within large-scale VSR systems remains limited. Rather than being treated as a structured geometric factor, head pose is often handled implicitly through preprocessing, overlooking its role in systematically altering articulation appearance and spatial relationships. Consequently, how pose-induced representation shifts affect sentence-level lip-reading performance remains poorly understood, particularly in unconstrained real-world settings.

\subsection{Conditional Feature Modulation and Representation Adaptation}

Conditional feature modulation techniques have demonstrated strong capability in adapting neural representations using auxiliary signals. Notably, FiLM \cite{perez_2018film} and related conditioning mechanisms such as conditional normalization \cite{dumoulin_2016} dynamically modulate intermediate feature maps to incorporate contextual or task-specific information. These methods have shown strong effectiveness in vision-language reasoning and multimodal learning tasks \cite{perez_2018film, devries_2017}, where conditioning signals guide the network to emphasize task-relevant representations while suppressing irrelevant variations.

Despite their success in representation adaptation, conditional modulation techniques remain underexplored in visual speech recognition. In particular, their potential to adapt visual features in response to geometric variations, such as those induced by head pose, has not been systematically investigated.

Meanwhile, modern VSR systems achieve strong performance on large-scale benchmarks such as LRS2 and LRS3 by leveraging powerful sequence modeling architectures and large-scale training. However, these approaches primarily rely on implicit handling of pose variation and do not explicitly model head-pose-dependent changes in visual articulation. Furthermore, existing pose-aware methods are often evaluated in controlled or synthetic settings, leaving their effectiveness under real-world pose variability insufficiently understood.

\noindent\textbf{Lack of Pose-Conditioned Feature Modulation in VSR.}
Taken together, these observations indicate that the role of head-pose variability in large-scale in-the-wild lip-reading remains underexplored. In particular, there is a lack of principled approaches that explicitly incorporate pose information to guide visual feature adaptation. This gap motivates the development of pose-conditioned modulation strategies for improving robustness to pose-induced distortions in visual speech recognition.

\section{Methodology}\label{method}

\begin{figure*}[t]
  \centering
   \includegraphics[scale=0.044]{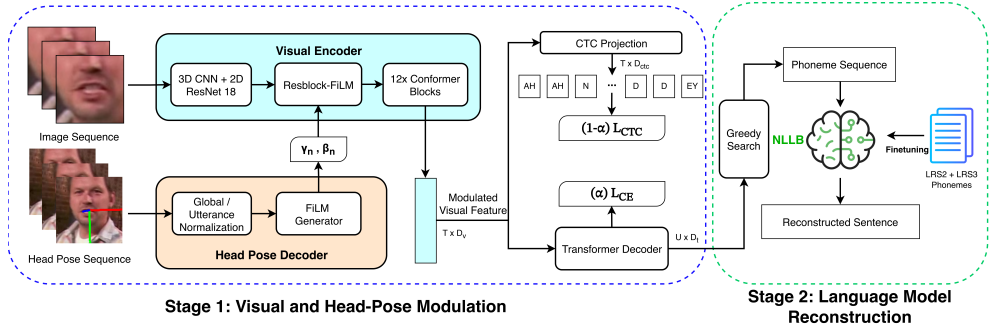} %0.48
%%have to change the caption.
   \caption{Overview of the proposed HP-VSR-ResFiLM framework. In Stage~1, visual features are extracted using a spatiotemporal visual encoder, while complementary geometric information is obtained from a head-pose encoder. The estimated head-pose representation is used to generate FiLM parameters that adaptively modulate intermediate visual representations through a residual FiLM block within the visual encoder, enabling pose-aware feature refinement. The modulated visual features are then processed by a temporal encoder and optimized using hybrid CTC/attention objectives for phoneme sequence prediction. In Stage~2, the predicted phoneme sequence is decoded using greedy autoregressive decoding and subsequently reconstructed into natural language using a fine-tuned NLLB model. Here, $T$ and $U$ denote the input and output sequence lengths, respectively; $D_v$ represents the visual feature dimension, $D_{ctc}$ denotes the CTC projection dimension, and $D_t$ corresponds to the phoneme vocabulary size.}
   
   \label{fig:overall_architecture}
\end{figure*}

The overall proposed system follows a two-stage phoneme-based architecture as shown in Figure~\ref{fig:overall_architecture}. In Stage~1, head-pose information is incorporated into the visual encoder through a residual FiLM block inserted after the 2D CNN frontend, enabling adaptive refinement of visual representations. The modulated features are subsequently processed by the Conformer encoder, followed by a Transformer decoder and optimized using a Connectionist Temporal Classification (CTC) objective for phoneme prediction.

In Stage 2, we conduct language-based reconstruction using a pre-trained large language model. Specifically, we adopt the same NLLB model as in prior work~\cite{teng_2026pvsr} and make no architectural modifications or additional fine-tuning. By keeping Stage 2 fixed, we ensure that performance differences are attributable solely to the proposed improvements in pose-aware visual feature modeling.

\subsection{Stage 1: Visual and Head-Pose Modulation.}
\textbf{Visual encoder.} We employ a visual backbone consisting of a spatiotemporal convolution frontend that uses a 3D convolutional layer with a stride of $1 \times 2 \times 2$ and a kernel size of $5 \times 7 \times 7$ to extract low- and mid-level spatiotemporal representations from input video frames. The frontend is followed by a modified ResNet-18~\cite{teng_2026pvsr} and a residual FiLM block that modulates visual features using head-pose information. Specifically, head-pose representations generated by the proposed head-pose encoder are injected into the residual block through FiLM, where pose-conditioned scaling and shifting parameters are learned to refine intermediate visual representations adaptively. This design enables the visual encoder to dynamically adjust its representations in response to head-pose variations while preserving the overall backbone architecture. The pose-aware visual features are subsequently processed by a 12-layer Conformer, which serves as the temporal back-end. The Conformer integrates multi-head self-attention and convolutional modules to capture both long-range temporal dependencies (e.g., phoneme coarticulation across frames) and local temporal dynamics (e.g., fine-grained lip movements), producing context-aware embeddings at each time step.

\noindent
\textbf{Pose-aware FiLM modulation.}
To enable head-pose-conditioned adaptation of visual features, we employ FiLM as a conditioning mechanism. FiLM applies a feature-wise affine transformation to intermediate activations of a neural network, allowing external pose information to influence visual representations adaptively.

Let $F_{i,c}$ denote the activation of the $c$-th feature channel corresponding to the $i$-th input sample. FiLM modulates this activation using a channel-wise scaling parameter $\gamma_{i,c}$ and shifting parameter $\beta_{i,c}$, defined as
\begin{equation}
\tilde{F}_{i,c} = \gamma_{i,c} F_{i,c} + \beta_{i,c}.
\label{eq:film}
\end{equation}

\noindent
\textbf{Head-pose FiLM generator.}
The normalized head-pose representation $\tilde{\mathbf{h}}_t$ is passed through a FiLM generator consisting of two fully connected layers with a ReLU activation to produce the pose-conditioned modulation parameters $\boldsymbol{\gamma}_t$ and $\boldsymbol{\beta}_t$. These parameters are subsequently applied to the intermediate visual feature maps according to Eq.~\eqref{eq:film}, enabling adaptive modulation of visual representations based on head-pose dynamics.

\noindent
\textbf{CTC~\cite{watanabe_IEEE_2017}.}
The projected encoder features are passed through a linear layer to produce phoneme-level logits. These logits are used to compute the CTC loss, which enforces monotonic alignment between input feature sequences and phoneme targets without requiring frame-level annotations. In the hybrid CTC/attention framework, the CTC objective complements the decoder loss, facilitating faster convergence and improving beam-search decoding performance.

\noindent
\textbf{Transformer decoder.}
The decoder architecture follows the standard Transformer formulation. Input tokens are first embedded and augmented with positional encodings. The embeddings are then processed by stacked decoder layers, each consisting of masked self-attention, encoder-decoder attention, and position-wise feed-forward networks with residual connections and dropout. A final linear projection maps decoder outputs to the phoneme vocabulary.

\noindent
\subsection{Stage 2: Language Model Reconstruction.}
For language reconstruction, we adopt the pre-trained NLLB model from~\cite{teng_2026pvsr}. The NLLB encoder transforms phoneme sequences into contextualized representations using multi-head self-attention and feed-forward layers. The decoder generates word tokens autoregressively by attending to encoder outputs and previously generated tokens. The final outputs are projected to the vocabulary space and normalized with softmax to produce word probabilities.

\section{Experiments}\label{experiment}

\subsection{Datasets}\label{dataset}

The LRS2 dataset \cite{son_2017lrs2} comprises 144,482 sentence-level video clips collected from BBC broadcasts, totaling approximately 224.5 hours of data. Each utterance contains a spoken sentence of up to 100 characters. The dataset is divided into four subsets: a pretraining set with 96,318 utterances (195 hours), a training set with 45,839 utterances (28 hours), a validation set with 1,082 utterances (0.6 hours), and a test set with 1,243 utterances (0.5 hours).

The LRS3 dataset \cite{afouras_2018lrs3} is the largest publicly available English audio-visual speech dataset, comprising 438.9 hours of video extracted from over 5,000 TED and TEDx talks. The dataset contains 151,819 utterances. Specifically, the pretraining set includes 118,516 utterances (408 hours), the train-validation set contains 31,982 utterances (30 hours), and the test set consists of 1,321 utterances (0.9 hours).

\subsection{Preprocessing}\label{preprocessing} 

The mouth region is cropped using a fixed $96 \times 96$ pixel bounding box. Each video frame is normalized by subtracting the mean and dividing by the standard deviation computed over the training set. To estimate head-pose information for each video sequence, including yaw, pitch, and roll angles, we employ the 6DRepNet model \cite{hempel_ICIP_2026} to extract per-frame head-pose values. For all sequences in the LRS2 and LRS3 datasets, head-pose parameters are computed frame by frame using this 6DRepNet estimator.

To account for dataset-specific pose biases and scale variations, we perform z-score normalization on the head-pose values. Specifically, we compute the dataset-level mean and standard deviation of yaw, pitch, and roll across all frames. Each frame-level head pose is then normalized by subtracting its mean and dividing by its standard deviation, yielding zero-mean, unit-variance representations. Let $\mathbf{h}_t = [y_t, p_t, r_t]$ denote the yaw, pitch, and roll angles at frame $t$, and let $\boldsymbol{\mu} = [\bar{y}, \bar{p}, \bar{r}]$ and $\boldsymbol{\sigma} = [\sigma_y, \sigma_p, \sigma_r]$ represent the dataset-level mean and standard deviation, respectively. The normalized head pose is computed as:
\begin{equation}
\tilde{\mathbf{h}}_t = \frac{\mathbf{h}_t - \boldsymbol{\mu}}{\boldsymbol{\sigma}}.
\end{equation}

This normalization strategy reduces inter-dataset bias and stabilizes the scale of head-pose variations across sequences, thereby facilitating more consistent learning when head-pose information is incorporated as an auxiliary cue. The normalized head-pose features are subsequently used in downstream processing without introducing any additional supervision. All normalization statistics are computed on the training split and applied consistently to validation and test sets.

For English grapheme-to-phoneme (G2P) conversion, text containing numbers, dates, and currency expressions is first normalized into their spoken forms using the text normalization module from the NVIDIA NeMo toolkit~\cite{zhang_2024nemo}. This preprocessing step ensures that the subsequent G2P conversion can handle non-standard written expressions. We then employ the open-source SoundChoice toolkit~\cite{ploujnikov_2022soundchoice}, which processes entire phrases rather than individual words, generating context-aware ARPAbet phoneme sequences. This preprocessing pipeline is consistent with our prior work~\cite{teng_2026pvsr} and adapted to the current experimental setting.

The resulting phoneme sequences are represented using the ARPAbet set, a standardized ASCII-based system for English phonemes, as summarized in Table~\ref{tab:phoneme_class_list}. Each symbol corresponds to a distinct sound and provides a computationally convenient alternative to the International Phonetic Alphabet (IPA). The representation includes both vowels (e.g., ``AA'', ``IY'', ``OW'') and consonants (e.g., ``K'', ``SH'', ``TH''), and is commonly used in automatic speech recognition and phoneme-based modeling \cite{teng_2026pvsr}. 

SentencePiece tokenization is used to convert the LRS2 and LRS3 transcriptions into phoneme sequences. The model is trained on 39 phoneme classes, excluding the special symbols ``\textless unk\textgreater'' and ``\_'', which represent unknown phonemes and spaces, respectively. This design is consistent with established phoneme-level AVSR settings~\cite{ma_2023avsr} and provides a balance between expressivity and computational efficiency.

\begin{table}[b]
  \centering
  \caption{Set of phoneme classes used in the proposed model, including vowel, consonant, and special tokens.}
  \label{tab:phoneme_class_list}
  \begin{tabular}{p{0.9\linewidth}}
    \toprule
    \{ \textless unk\textgreater, \_, AA, AE, AH, AO, AW, AY, B, CH, D, DH, EH, ER, EY, F, G, HH, IH, IY, JH, K, L, M, N, NG, OW, OY, P, R, S, SH, T, TH, UH, UW, V, W, Y, Z, ZH \} \\
    \bottomrule
  \end{tabular}
\end{table}

\subsection{Head-Pose Statistics}\label{headpose_statistic}
%%%% need to show pitch and roll frames data in the table, them say why vsr only use yaw. face alignment centering the face can cause roll and pitch to be 0˚?

% \begin{figure*}[tb]
%     \centering
%     % --- LRS2 + LRS3 Histograms ---
%     \includegraphics[scale=0.07]{figures/LRS2_headpose_histograms.eps} \quad
%     \includegraphics[scale=0.07]{figures/LRS3_headpose_histograms.eps}

%     \caption{Comparison of LRS2 and LRS3 yaw distributions. Left: LRS2; Right: LRS3.}
%     \label{fig:histograms_comparison}
% \end{figure*}
% Figure~\ref{fig:histograms_comparison} shows the histogram distributions of yaw angles, where both datasets exhibit a strong bias toward near-frontal views, with a pronounced peak around $0^\circ$. 
% Despite this concentration, the distributions display a clear long-tail behavior, indicating the presence of a substantial number of semi-profile and profile frames.

% Figures~\ref{fig:heatmaps_comparison} and \ref{fig:LRS2_LRS3_yaw_comparison} illustrate the head-pose characteristics of the LRS2 and LRS3 datasets. 

To better understand the extent of head-pose variation in large-scale VSR datasets, we analyze the distribution of head poses in LRS2 and LRS3. Specifically, we examine the joint distribution of yaw and pitch angles, as well as the marginal distribution of yaw, to characterize the prevalence of non-frontal views. 

\begin{figure*}[tb]
    \centering
    % --- Heatmaps ---
    \begin{subfigure}[b]{0.48\textwidth}
        \centering
        \includegraphics[scale=0.08]{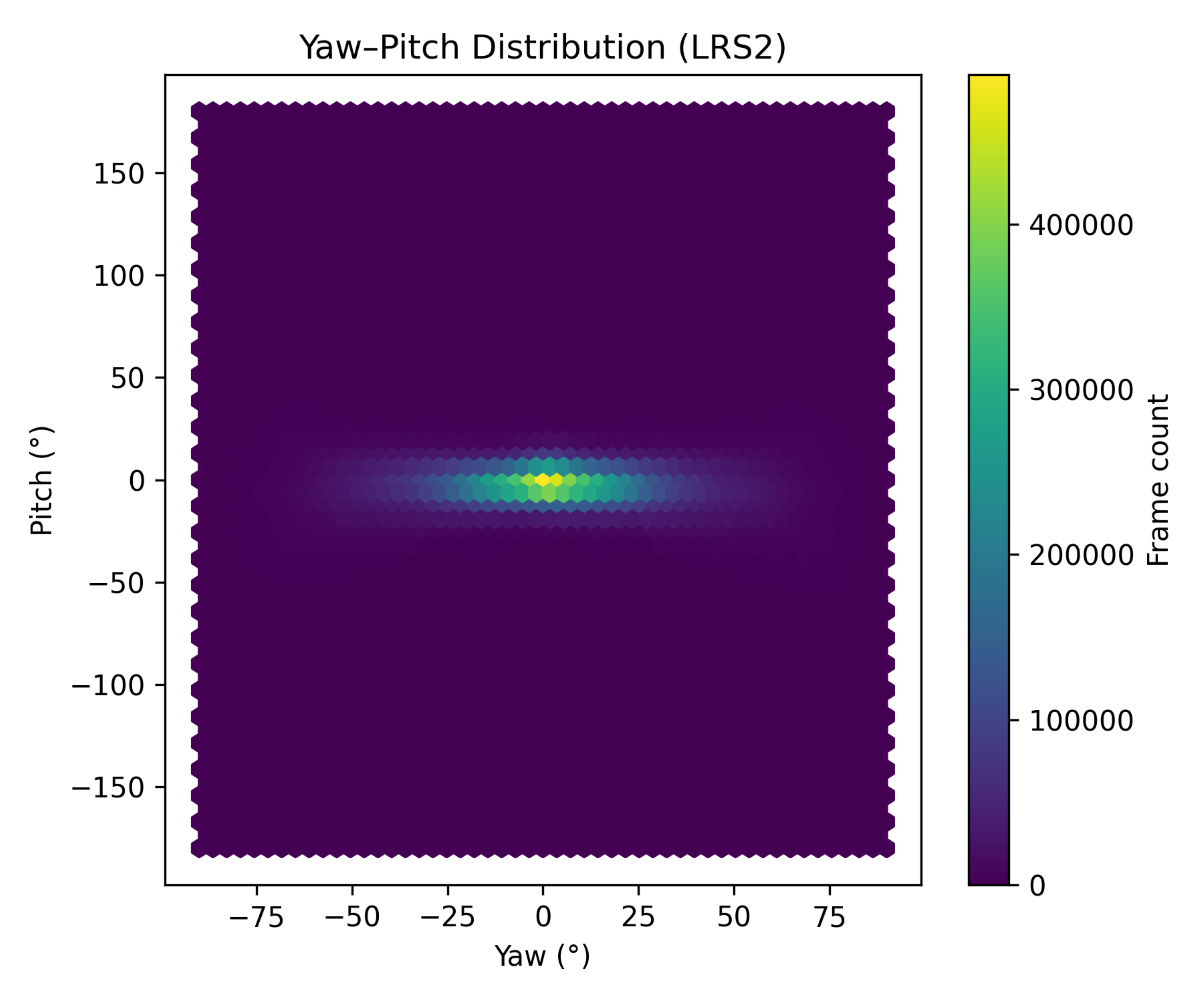}
        \caption{LRS2 Yaw-Pitch Heatmap}
        \label{fig:lrs2_heatmap}
    \end{subfigure}%
    \hfill
    \begin{subfigure}[b]{0.48\textwidth}
        \centering
        \includegraphics[scale=0.08]{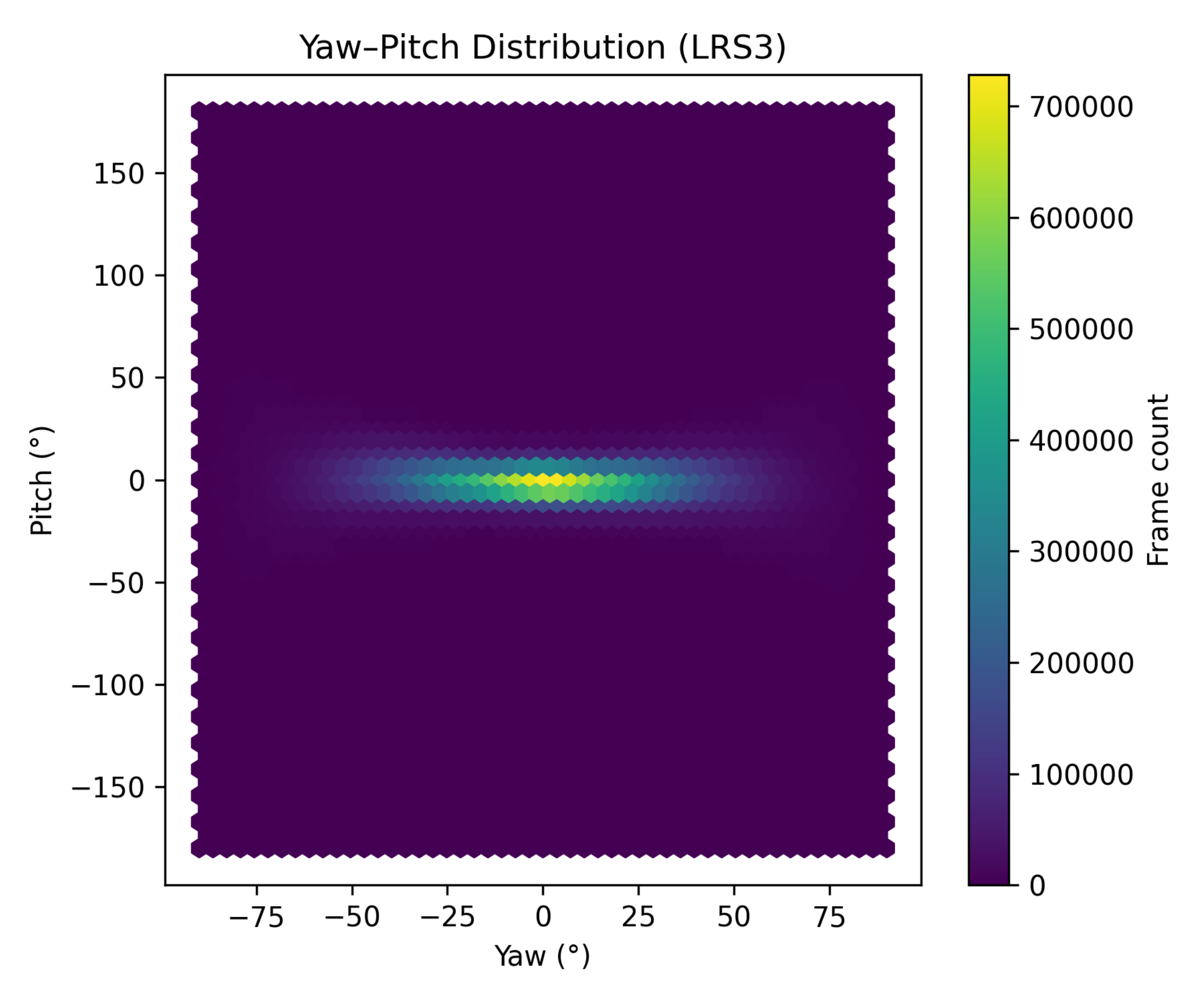}
        \caption{LRS3 Yaw-Pitch Heatmap}
        \label{fig:lrs3_heatmap}
    \end{subfigure}
    \caption{Yaw–pitch distribution heatmaps for the LRS2 and LRS3 datasets. The left panel shows LRS2, while the right panel shows LRS3. Both datasets are dominated by frontal and semi-frontal views, with a non-negligible presence of large out-of-plane rotations.}
    \label{fig:heatmaps_comparison}
\end{figure*}

Figure~\ref{fig:heatmaps_comparison} presents the joint yaw-pitch distributions for LRS2 and LRS3 as heatmaps. While both datasets exhibit a high-density region around frontal orientations, the distributions extend broadly across the yaw-pitch space rather than being tightly concentrated. This widespread dispersion indicates that a substantial portion of frames involve non-frontal views with noticeable out-of-plane head rotations, reflecting the unconstrained nature of in-the-wild video recordings.

\begin{figure*}[tb]
  \centering
   \includegraphics[scale=0.08]{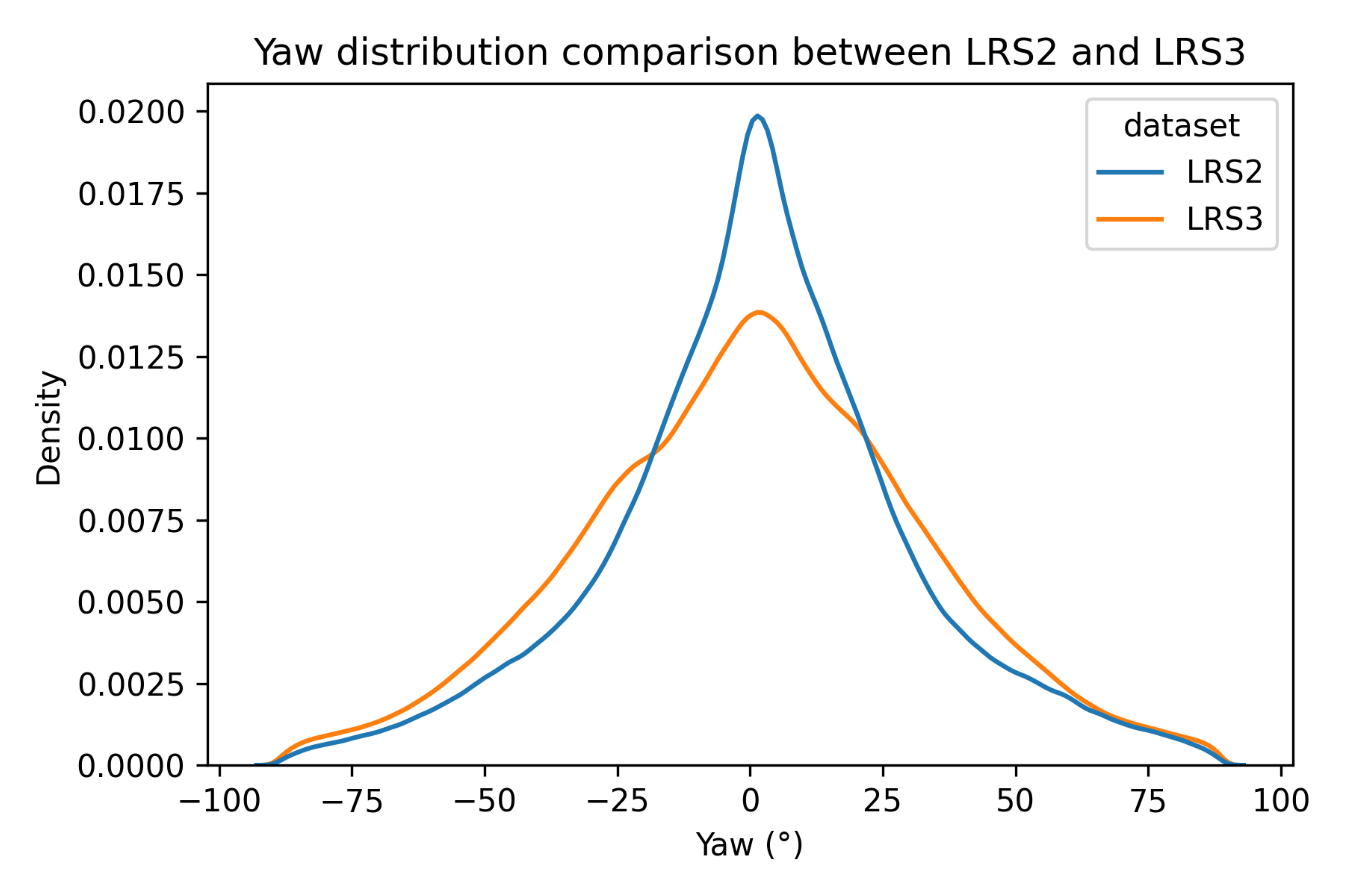} %0.48
   \caption{Comparison of yaw angle distributions in LRS2 and LRS3. Both datasets show a strong concentration around near-frontal poses, with a gradual decrease toward larger yaw angles, indicating pose variability in unconstrained settings.}
   \label{fig:LRS2_LRS3_yaw_comparison}
\end{figure*}

Figure~\ref{fig:LRS2_LRS3_yaw_comparison} further compares the marginal yaw distributions of the two datasets using density curves. Although both distributions show a peak around frontal orientations, they exhibit long tails and significant spread toward larger yaw angles, indicating that non-frontal views occur frequently. In particular, LRS3 demonstrates a noticeably wider distribution, suggesting greater pose variability. This difference is consistent with the TED Talk format, where speakers often turn their heads to engage different sections of the audience.

Table~\ref{tab:headpose_distribution} reports the distribution of frames across yaw, pitch, and roll angle ranges for the LRS2 and LRS3 datasets. Although near-frontal views (within $15^\circ$) constitute the largest portion of the data, a substantial proportion of frames exhibit pronounced yaw variations. In particular, frames with yaw angles exceeding $30^\circ$ account for $26.51\%$ of LRS2 and $34.78\%$ of LRS3, indicating that more than one-quarter to one-third of the data involves significant non-frontal views. Furthermore, extreme poses beyond $45^\circ$ still represent $13.31\%$ and $16.89\%$ of the datasets, respectively, which is non-negligible for reliable lip-reading. These results highlight that, despite a concentration around frontal orientations, yaw variation is both widespread and substantial, introducing frequent self-occlusion and geometric distortion in the lip region. This level of pose variation is more pronounced than typically assumed and underscores the need for VSR models to handle non-frontal views explicitly.

\begin{table}[b]
\centering
\caption{Yaw angle distribution statistics for the LRS2 and LRS3 datasets. More than 50\% of frames in both datasets exhibit yaw angles greater than $15^\circ$, indicating the presence of substantial non-frontal head motion.}
\label{tab:headpose_distribution}
\begin{tabular}{lcccc}
\toprule
\multirow{2}{*}{\textbf{Yaw Range}} & \multicolumn{2}{c}{\textbf{LRS2}} & \multicolumn{2}{c}{\textbf{LRS3}} \\
\cmidrule(lr){2-3} \cmidrule(lr){4-5}
 & \textbf{Frames} & \textbf{Percentage (\%)} & \textbf{Frames} & \textbf{Percentage (\%)} \\
\midrule
$<15^\circ$ (Frontal) & 9{,}450{,}773 & 46.99 & 14{,}629{,}800 & 37.25 \\
$15$--$30^\circ$ (Semi-frontal) & 5{,}327{,}743 & 26.49 & 10{,}988{,}773 & 27.98 \\
$30$--$45^\circ$ (Side) & 2{,}654{,}236 & 13.20 & 7{,}025{,}340 & 17.89 \\
$45$--$60^\circ$ (Profile/Extreme) & 1{,}526{,}894 & 7.59 & 3{,}889{,}761 & 9.90 \\
$>60^\circ$ (Extra Extreme) & 1{,}150{,}859 & 5.72 & 2{,}743{,}914 & 6.99 \\
\bottomrule
\end{tabular}
\end{table}

In contrast, pitch and roll angles are highly concentrated around near-frontal orientations. For pitch, $85.16\%$ and $87.36\%$ of frames fall within $15^\circ$ in LRS2 and LRS3, respectively, while roll exhibits a similar trend with $83.52\%$ and $88.56\%$ of frames in the same range. Only a small fraction of frames exceed $30^\circ$ for both pitch and roll in either dataset. These head-pose statistics are computed from raw video frames before face alignment. The face-alignment preprocessing further normalizes vertical (pitch) and in-plane (roll) rotations toward canonical frontal views. As a result, the effective variation of pitch and roll observed by the model during training and inference is likely even smaller than indicated by these statistics. In contrast, yaw variations are less constrained by alignment and remain substantially more pronounced, making them a primary source of pose-induced difficulty in visual speech recognition.

Although prior work~\cite{ma_2023avsr,teng_2026pvsr,yeo_2024} recognizes that LRS2 and LRS3 contain significant pose variation, existing studies typically treat this variability implicitly and do not report quantitative head-pose statistics. By explicitly analyzing yaw and pitch distributions and pose-dependent frame counts, our study reveals that non-frontal views constitute a non-trivial portion of both datasets. This analysis directly motivates our head-pose–aware modeling strategy, which addresses pose-induced appearance changes and self-occlusion in lip-reading.

\subsection{Loss Function}\label{loss_function} %need to rephrase to diff from PVSR archiv paper.
To simplify the conditional independent assumption of CTC and enforce monotonic alignments, we use a hybrid CTC/Attention architecture \cite{watanabe_IEEE_2017} in this study. The following is the definition of the overall training goal:
\begin{equation}
\mathcal{L} = \alpha \mathcal{L}_\text{CE} + (1 - \alpha) \mathcal{L}_\text{CTC}
\label{eq:hybrid_loss}
\end{equation}

\noindent
where $\mathcal{L}_\text{CTC}$ is the CTC loss and $\mathcal{L}_\text{CE}$ is the cross-entropy loss. The adjustable parameter $\alpha \in [0,1]$ controls the trade-off between the cross-entropy loss and the CTC loss.

\subsection{Experimental Setup}\label{setup}

All experiments are conducted on a single NVIDIA A6000 GPU with 49GB of memory. For Stage~1, following the experimental protocol of~\cite{teng_2026pvsr}, we train the model for 80 epochs using the AdamW optimizer with an initial learning rate of $1 \times 10^{-4}$. We use a cosine learning-rate scheduler with a 5-epoch warm-up. The maximum number of frames per training batch is set to 1{,}800. We also initialize the visual encoder from the publicly released Auto-AVSR model~\cite{ma_2023avsr}, which is trained on a large-scale VSR dataset comprising LRW~\cite{chung_2017lrw}, LRS2, LRS3, VoxCeleb2~\cite{chung_2018_voxceleb2}, and AVSpeech~\cite{ephrat_2018_avspeech} with automatically generated transcriptions, totaling 3,448 hours. In contrast to the original training protocol, we fine-tune the model using only the labeled data from LRS2 or LRS3, without incorporating additional external data during training.

The 3D convolution frontend of the visual encoder is kept frozen, while the remaining components are fine-tuned. This training protocol ensures a controlled comparison across ablation variants by avoiding confounding effects from differing fine-tuning strategies. Early stopping with a patience of 5 epochs is applied to mitigate overfitting. For evaluation, we report results by averaging the model parameters from the final 10 checkpoints, which stabilizes performance estimates and reduces variance from epoch-to-epoch fluctuations. 

For Stage~2, we reimplement the LLM architecture and training configuration introduced in our prior work~\cite{teng_2026pvsr}, and do not perform any additional fine-tuning in this study. In~\cite{teng_2026pvsr}, the model was fine-tuned using the AdamW optimizer with an initial learning rate of $5 \times 10^{-5}$. The default English tokenizer is employed, and phoneme sequences are treated as a distinct input language, enabling the model to learn a mapping from phonemic representations to natural language. During decoding, beam search is applied, with hyperparameters tuned on the validation set; a beam size of 15 and a length penalty of 0.1 were found to yield the best performance.

\subsection{Evaluation Metrics}\label{metrics}

To evaluate lip-reading performance, we adopt WER \cite{jelinek_1975wer} as the primary metric. WER quantifies the discrepancy between the predicted text and the reference transcript by accounting for substitutions, deletions, and insertions. Formally, WER is defined as:

\begin{equation}
\text{WER} = \frac{S + D + I}{N}
\label{eq:wer}
\end{equation}

\noindent
where \(S\), \(D\), and \(I\) denote the number of substitutions, deletions, and insertions, respectively, and \(N\) represents the total number of words in the reference transcript. 

In addition to WER, we report phoneme error rate (PER) to evaluate the accuracy of phoneme-level predictions. Together, these metrics provide a comprehensive and fine-grained assessment of lip-reading performance.

\subsection{Comparison with State-of-the-Art Methods}

\begin{table*}[b]
\centering
\setlength{\tabcolsep}{3pt}
\caption{Comparison of phoneme-based and word-based visual speech recognition methods on the LRS2 and LRS3 datasets. The table reports input modality, LLM usage, external data usage, and target-domain training duration. Training Data (h) denotes the amount of labeled LRS2/LRS3 data used during supervised fine-tuning and evaluation. Extra Data indicates whether additional datasets are used during pretraining or training beyond the target benchmarks. Our methods are initialized from Auto-AVSR pretrained weights trained on 3,448 hours of large-scale audio-visual data. Still, they are fine-tuned only on the standard LRS2/LRS3 training sets without additional external data during adaptation. Similarly, the Auto-AVSR Fine-tuned Baseline is initialized from the same pretrained weights and fine-tuned only on the target datasets. V-ASR and PV-ASR are re-evaluated using greedy autoregressive decoding for consistency with the proposed framework.}
% \renewcommand{\arraystretch}{0.95}
% \small
\label{tab:sota_lrs2lrs3}
\resizebox{\linewidth}{!}{
\begin{tabular}{lccccccccc}
\toprule
 \multirow{2}{*}{\textbf{Method}} & \multirow{2}{*}{\makecell{\textbf{Phoneme} \\ \textbf{-based}}} & \multirow{2}{*}{\textbf{Input}} & \multirow{2}{*}{\textbf{LLM}} & \multirow{2}{*}{\makecell{\textbf{Extra} \\ \textbf{Data}}} & 
\multicolumn{2}{c}{\textbf{LRS2}} & \multicolumn{2}{c}{\textbf{LRS3}} \\
\cmidrule(lr){6-7} \cmidrule(lr){8-9}
 & & & & & \textbf{Training} & \textbf{WER} & \textbf{Training} & \textbf{WER} \\
 & & & & & \textbf{Data [h]} & \textbf{[\%]} & \textbf{Data [h]} & \textbf{[\%]} \\
\midrule
ASSTGCN \cite{sheng_2021asstgcn} & - & Video+38-Points & - & - & 223 & 55.7 & 438 & 62.7 \\
Hyb.-Conf. \cite{ma_IEEE_2021} & - & Video & - & \cmark & 381 & 37.9 & 590 & 43.3 \\
VTP \cite{prajwal_2022subword} & - & Video & - & \cmark & 698 & 48.9 & 698 & 40.6 \\
CM-aux \cite{ma_2022multilanguage} & - & Video & - & - & 223 & 32.9 & 438 & 36.3 \\
VTP \cite{prajwal_2022subword} & - & Video & - & \cmark & 2676 & 22.6 & 2676 & 30.7 \\
Auto-AVSR \cite{ma_2023avsr} & - & Video & - & \cmark & 818 & 27.9 & 818 & 33.0 \\
Auto-AVSR \cite{ma_2023avsr} & - & Video & - & \cmark & 3448 & 14.6 & 3448 & 19.1 \\
VALLR \cite{thomas_2025vallr} & \cmark & Video & LLaMA & - & 28 & 20.8 & 30 & 18.7 \\
ViT-3D \cite{serdyuk_2022} & - & Video & - & \cmark & - & - & 90000 & 17.0 \\
SyncVSR \cite{ahn_2024syncvsr} & \cmark & Video & - & - & 223 & 28.9 & 438 & 31.2 \\
GLip \cite{wang_2025glip} & \cmark & Video & - & - & 223 & 27.4 & 438 & 30.1 \\
V-ASR \cite{teng_2026pvsr} & \cmark & Video & NLLB(1.3B) & - & 223 & 35.2 & 438 & 41.2 \\
PV-ASR \cite{teng_2026pvsr} & \cmark & Video+117-Points & NLLB(1.3B) & - & 223 & 26.9 & 438 & 38.7 \\
Auto-AVSR Finetuned Baseline & - & Video & - & - & 223 & 33.0 & 438 & 29.8 \\ 
HP-VSR-Base (Our) & \cmark & Video+Head Pose & NLLB(1.3B) & - & 223 & 33.3 & 438 & 40.8 \\
HP-VSR-FiLMFuse(L4) (Our) & \cmark & Video+Head Pose & NLLB(1.3B) & - & 223 & 23.3 & 438 & 33.0 \\
HP-VSR-ResFiLM (Our) & \cmark & Video+Head Pose & NLLB(1.3B) & - & 223 & 24.7 & 438 & 30.3 \\
\bottomrule
\end{tabular}
}
\end{table*}

Table~\ref{tab:sota_lrs2lrs3} compares the proposed method with recent word-based and phoneme-based visual speech recognition approaches on the LRS2 and LRS3 benchmarks. The comparison additionally reports input modality, language model usage, and the amount of external training data to provide a fair assessment across different experimental settings. We report results for three variants of our framework: HP-VSR-Base, which directly incorporates head-pose information through feature fusion; HP-VSR-FiLMFuse, which applies FiLM-based modulation with explicit pose fusion at deeper visual layers; and HP-VSR-ResFiLM, the proposed residual FiLM formulation.

% Existing visual-only lip-reading approaches~\cite{sheng_2021asstgcn,ma_IEEE_2021} rely primarily on appearance-based visual representations without explicit pose-aware conditioning, making them sensitive to pose-induced geometric distortion and self-occlusion under unconstrained viewing conditions. Although large-scale pretraining~\cite{prajwal_2022subword,ma_2023avsr} substantially improves recognition performance, these gains rely heavily on massive external audio-visual corpora and computationally intensive pretraining pipelines. More recently, phoneme-based frameworks~\cite{thomas_2025vallr,teng_2026pvsr} have explored decoupling visual recognition from linguistic reconstruction through external language models, enabling improved phonetic modeling and more flexible integration of linguistic priors.

Existing visual-only lip-reading approaches~\cite{sheng_2021asstgcn,ma_IEEE_2021} rely primarily on appearance-based visual representations without explicit pose-aware conditioning, making them sensitive to pose-induced geometric distortion and self-occlusion under unconstrained viewing conditions. Although large-scale pretraining~\cite{prajwal_2022subword,ma_2023avsr} substantially improves recognition performance, these gains rely heavily on massive external audio-visual corpora and computationally intensive pretraining pipelines. More recently, phoneme-based frameworks~\cite{thomas_2025vallr,teng_2026pvsr} have explored decoupling visual recognition from linguistic reconstruction through external language models, enabling improved phonetic modeling and more flexible integration of linguistic priors. Recent phoneme-based approaches have further investigated alternative strategies for improving VSR robustness. SyncVSR~\cite{ahn_2024syncvsr} introduces crossmodal audio-token synchronization as an auxiliary learning objective to provide frame-level supervision, while GLip~\cite{wang_2025glip} employs a progressive global--local visual representation learning framework to improve recognition under challenging visual conditions. These approaches demonstrate the effectiveness of additional supervisory signals and enhanced visual representations for phoneme-based VSR. However, they do not explicitly incorporate head-pose information as a conditioning signal during visual feature extraction. In contrast, our HP-VSR framework explicitly models head-pose dynamics and uses them to adaptively modulate visual representations, thereby targeting pose-induced variations directly.

To establish a controlled comparison under limited target-domain supervision, we additionally fine-tune the released Auto-AVSR pretrained model using only the standard LRS2 and LRS3 training sets without introducing additional external data during adaptation. The resulting performance gap compared with the original large-scale Auto-AVSR results highlights the substantial dependence of end-to-end visual speech recognition systems on large-scale pretraining data.

In addition, for fair comparison under practical inference settings, the reported V-ASR and PV-ASR results are re-evaluated using greedy autoregressive decoding rather than the teacher-forced decoding protocol reported in the original work~\cite{teng_2026pvsr}. Teacher-forced evaluation conditions the decoder on ground-truth previous tokens during inference and may therefore overestimate real-world recognition performance. In contrast, greedy decoding reflects fully autoregressive inference and provides a more realistic assessment of end-to-end transcription robustness.

Under these controlled evaluation settings, HP-VSR-Base, which incorporates head-pose information through direct feature fusion, achieves performance comparable to the video-only baseline but does not consistently improve robustness across datasets. This observation suggests that naive multimodal fusion alone is insufficient for effectively utilizing pose information within the visual encoder. In contrast, both HP-VSR-FiLMFuse (L4) and the proposed HP-VSR-ResFiLM achieve substantial improvements over the baseline approaches, demonstrating the effectiveness of pose-conditioned feature modulation. 

As shown in Table~\ref{tab:sota_lrs2lrs3}, the proposed HP-VSR-ResFiLM achieves WERs of 24.7\% on LRS2 and 30.3\% on LRS3 while being fine-tuned only on the standard LRS2/LRS3 training sets using pretrained Auto-AVSR initialization. Compared with the video-only V-ASR baseline~\cite{teng_2026pvsr}, the proposed method reduces WER from 35.2\% to 24.7\% on LRS2 and from 41.2\% to 30.3\% on LRS3, demonstrating that explicit pose-conditioned modulation consistently improves phoneme-based visual speech recognition across both datasets. Furthermore, compared with PV-ASR~\cite{teng_2026pvsr}, which relies on dense 117-point facial landmark representations, HP-VSR-ResFiLM achieves lower WER on both datasets while requiring only head-pose information in addition to the video stream.

Among recent phoneme-based approaches, SyncVSR~\cite{ahn_2024syncvsr} and GLip~\cite{wang_2025glip} provide strong recent baselines. HP-VSR-ResFiLM achieves a WER of 24.7\% on LRS2, outperforming both SyncVSR (28.9\%) and GLip (27.4\%). On LRS3, HP-VSR-ResFiLM achieves 30.3\%, improving over SyncVSR (31.2\%) while remaining competitive with GLip (30.1\%). These results demonstrate that pose-conditioned modulation provides competitive performance with recent phoneme-based VSR approaches without introducing additional external data during target-domain adaptation.

Although the Auto-AVSR Fine-tuned Baseline achieves a lower WER on LRS3, it remains a fully end-to-end sequence generation framework directly optimized for word-level transcription. In contrast, the proposed approach explicitly decomposes visual speech recognition into phoneme prediction and language-level reconstruction. While this intermediate phoneme representation may introduce additional reconstruction errors, it provides improved interpretability, modularity, and flexibility for analyzing pose-aware visual speech modeling. These results suggest that the proposed FiLM-based modulation primarily improves the robustness of phoneme-level visual representations rather than directly optimizing end-to-end language reconstruction performance.

The results further reveal important differences between fusion-based and modulation-based integration strategies. While direct feature fusion introduces additional pose features at a later stage, FiLM-based conditioning adaptively modulates intermediate visual representations according to pose dynamics during feature extraction. This enables the network to recalibrate visual representations according to head orientation and results in improved robustness across the two benchmarks. HP-VSR-FiLMFuse (L4) achieves the strongest performance on LRS2, indicating that explicit pose fusion can effectively complement deep FiLM conditioning. In contrast, the proposed HP-VSR-ResFiLM achieves the strongest performance among our pose-aware variants on LRS3 while maintaining competitive performance on LRS2, providing the most balanced performance across both benchmarks. These findings suggest that residual FiLM conditioning provides a robust mechanism for pose-aware adaptation, while explicit feature fusion can provide additional gains when combined with carefully selected deep-layer modulation.

\subsection{Ablation Studies}\label{results}

\subsubsection{Impact of FiLM Layer Placement and Multimodal Head-pose Fusion}

We conduct a systematic ablation study under strictly controlled experimental settings to analyze the contribution of FiLM-based pose modulation and its interaction with multimodal head-pose fusion. All variants share identical backbone architecture, optimizer configuration, training schedule, preprocessing pipeline, hyperparameter settings, random seeds, cross-validation splits, and decoding framework, differing only in the placement of FiLM modulation and the inclusion of head-pose fusion. In addition, Stage~2 is fixed across all experiments to ensure that performance differences arise solely from architectural modifications in Stage~1. All experiments are evaluated using identical protocols on both LRS2 and LRS3, enabling reproducible and fair comparisons while isolating the effect of pose-conditioned modulation from other confounding factors.

Table~\ref{tab:ablation_film_pose} summarizes the ablation results. Incorporating head-pose information through FiLM-based conditioning consistently improves both phoneme-level recognition and downstream sentence reconstruction compared with the unmodulated HP-VSR-Base across both datasets, confirming that pose cues provide complementary information beyond visual speech features alone. The proposed HP-VSR-ResFiLM achieves the strongest overall cross-dataset performance, obtaining 16.20\% / 20.96\% PER and 24.68\% / 30.32\% WER on LRS2 and LRS3, respectively. Relative to HP-VSR-Base, this corresponds to absolute WER reductions of 8.62\% on LRS2 and 10.51\% on LRS3. HP-VSR-FiLMFuse (L4) performance degrades on LRS3, indicating reduced cross-dataset robustness. Overall, the results demonstrate that pose-conditioned feature adaptation substantially improves recognition under unconstrained pose variations while highlighting the importance of modulation strategy and placement.

\begin{table*}[tb]
\centering
\caption{Ablation study on FiLM modulation strategies and head-pose fusion. Stage~1 produces phoneme-level predictions evaluated using PER, while Stage~2 generates word-level outputs evaluated using WER. Stage~2 is fixed across all experiments to ensure that performance differences arise solely from architectural variations in Stage~1. The results show that incorporating a single FiLM-modulated residual block after the visual frontend consistently provides the best overall performance, demonstrating the effectiveness of simple pose-conditioned feature modulation. Additional experiments on deeper FiLM placement further analyze the effect of modulation depth and its interaction with multimodal fusion.}
\label{tab:ablation_film_pose}
\resizebox{\linewidth}{!}{
\begin{tabular}{l c c c c c c}
\toprule
\textbf{Configuration} & \multirow{2}{*}{\makecell{\textbf{FiLM} \\ \textbf{Placement}}} & \textbf{Fusion} 
& \multicolumn{2}{c}{\textbf{Stage 1 (PER, \%)}} 
& \multicolumn{2}{c}{\textbf{Stage 2 (WER, \%)}} \\
\cmidrule(lr){4-5} \cmidrule(lr){6-7}
 &  &  & \textbf{LRS2} & \textbf{LRS3} & \textbf{LRS2} & \textbf{LRS3} \\
\midrule
PV-ASR~\cite{teng_2026pvsr} & -- & \cmark   & 18.53 & 25.25 & 26.55 & 35.96 \\
V-ASR~\cite{teng_2026pvsr} & -- & --        & 23.26 & 26.90 & 34.91 & 38.62 \\
HP-VSR-Base & -- & \cmark                   & 23.28 & 29.62 & 33.30 & 40.83 \\
HP-VSR-FiLM (L1--2) & Early & --            & 18.98 & 34.11 & 29.16 & 48.08 \\
HP-VSR-FiLM (L3--4) & Late & --             & 21.49 & 47.56 & 32.68 & 66.59 \\
HP-VSR-FiLM (L4) & Deep & --                & 16.31 & 22.62 & 24.69 & 32.75 \\
HP-VSR-FiLMFuse (L1--2) & Early & \cmark    & 19.42 & 31.79 & 27.85 & 43.51 \\
HP-VSR-FiLMFuse (L3--4) & Late & \cmark     & 17.28 & 22.92 & 25.44 & 32.54 \\
HP-VSR-FiLMFuse (L4) & Deep & \cmark        & \textbf{15.81} & 22.85 & \textbf{23.25} & 33.00  \\
Global FiLM & Post-ResNet & --              & 16.21 & 21.29 & 24.58 & 30.43 \\
HP-VSR-ResFiLM & Post-ResNet & --           & 16.20 & \textbf{20.96} & 24.68 & \textbf{30.32} \\
HP-VSR-ResFiLMFuse & Post-ResNet & \cmark   & 19.08 & 26.19 & 27.06 & 36.68 \\
\bottomrule
\end{tabular}
}
\end{table*}

The effectiveness of FiLM modulation is strongly dependent on its placement within the visual hierarchy. Applying FiLM to early ResNet18 layers, as implemented in HP-VSR-FiLM (L1--2), significantly degrades performance, particularly on LRS3, where WER increases from 40.83\% to 48.08\%. This behavior suggests that low-level convolutional representations primarily encode local texture and edge statistics that remain insufficiently discriminative for stable pose-conditioned modulation. In contrast, deeper representations contain higher-level articulatory and semantic information that are more suitable for adaptive conditioning. However, directly modulating multiple deep layers without residual stabilization produces severe optimization instability. In particular, HP-VSR-FiLM (L3--4) suffers substantial degradation on LRS3, reaching 66.59\% WER, indicating that aggressive modulation of latent visual representations can amplify pose-induced feature shifts and destabilize temporal alignment learning.

We further analyze the interaction between FiLM conditioning and explicit head-pose fusion. The effect of fusion depends strongly on the FiLM placement strategy. For early-layer modulation, HP-VSR-FiLMFuse (L1--2) improves upon HP-VSR-FiLM (L1--2), reducing WER from 29.16\% / 48.08\% to 27.85\% / 43.51\% on LRS2 and LRS3, respectively. More substantial gains are observed for deeper modulation strategies. In particular, HP-VSR-FiLMFuse (L3--4) dramatically improves LRS3 performance, reducing WER from 66.59\% to 32.54\%, while simultaneously lowering PER from 47.56\% to 22.92\%. Similarly, HP-VSR-FiLMFuse (L4) achieves the lowest LRS2 WER among all evaluated variants, improving from 24.69\% to 23.25\%. These consistent improvements indicate that explicit head-pose fusion can effectively stabilize FiLM conditioning, particularly when modulation is applied to deeper visual representations where pose-induced feature shifts become more pronounced.

The proposed residual FiLM formulation further demonstrates advantages over alternative modulation strategies. Compared with Global FiLM, HP-VSR-ResFiLM achieves lower PER on both datasets and obtains the best WER on LRS3 while maintaining competitive performance on LRS2. Although Global FiLM achieves a marginally lower WER on LRS2 (24.58\% versus 24.68\%), its higher PER and weaker gains on LRS3 suggest that globally conditioning the entire feature representation is less effective than localized residual adaptation. These results support the hypothesis that residual conditioning preserves the underlying visual feature manifold while enabling pose-aware refinement through additive modulation, thereby improving optimization stability and cross-dataset generalization.

Nevertheless, despite the benefits of fusion for deeper FiLM configurations, HP-VSR-ResFiLM remains the most balanced and robust approach across datasets. HP-VSR-ResFiLM achieves the best performance on LRS3 while maintaining competitive results on LRS2 without requiring explicit pose fusion. These findings suggest that residual FiLM conditioning provides an effective mechanism for integrating pose information through adaptive feature refinement. In contrast, explicit fusion primarily serves as a complementary stabilization strategy for more aggressively modulated architectures.

Overall, the ablation study demonstrates that the primary advantage of FiLM lies in enabling adaptive pose-conditioned refinement of visual representations rather than simply increasing model capacity. The observed performance trends remain consistent across both LRS2 and LRS3, indicating that the improvements are not dataset-specific artifacts but reflect stable architectural behavior under varying pose distributions. While several variants achieve strong results on individual datasets, their performance varies considerably across evaluation conditions. HP-VSR-FiLMFuse (L4) achieves the best LRS2 performance. In contrast, HP-VSR-ResFiLM consistently delivers strong PER and WER performance across both datasets and achieves the best results on the more challenging LRS3 benchmark. Furthermore, the substantial degradation observed in HP-VSR-FiLM (L3--4) highlights the sensitivity of direct deep-layer modulation, whereas the residual formulation maintains stable optimization and effective pose adaptation. These findings support the design choice of introducing a single residual FiLM block after the visual frontend, where pose-aware conditioning can refine high-level visual representations while preserving stable feature propagation throughout the network.

\subsubsection{Performance under Head-Pose Variations}

We evaluate robustness to pose variation by grouping test samples according to yaw angle, following~\cite{cheng_2020icassp}. For each sample, frame-level yaw angles are averaged to obtain a scene-level estimate, and samples are categorized into three ranges: near-frontal ($<15^\circ$), moderate ($15^\circ$--$30^\circ$), and large ($>30^\circ$) poses. We focus on yaw as it represents the dominant source of pose variation in both LRS2 and LRS3, while pitch and roll remain largely within near-frontal ranges.

Table~\ref{tab:pose_robustness} reports performance across different pose conditions. Overall, all methods experience degradation as yaw angle increases, highlighting the inherent difficulty of visual speech recognition under large out-of-plane head rotations. However, the magnitude of degradation differs substantially across architectures, providing important insight into the effectiveness of explicit pose modeling and feature modulation strategies.

\begin{table*}[tb]
\centering
\caption{Performance evaluation under different head-pose ranges. Test samples are grouped according to yaw angle to analyze the effect of pose variation. Stage~1 reports PER, while Stage~2 reports WER. Lower values indicate better performance.}
\label{tab:pose_robustness}
\resizebox{\linewidth}{!}{
\begin{tabular}{l c c c c c}
\toprule
\textbf{Method} &
\multirow{2}{*}{\makecell{\textbf{Yaw} \\ \textbf{Angles}}}
&
\multicolumn{2}{c}{\textbf{Stage 1 (PER, \%)}} &
\multicolumn{2}{c}{\textbf{Stage 2 (WER, \%)}} \\
\cmidrule(lr){3-4}
\cmidrule(lr){5-6}
&
& \textbf{LRS2}
& \textbf{LRS3}
& \textbf{LRS2}
& \textbf{LRS3} \\
\midrule

\multirow{3}{*}{V-ASR~\cite{teng_2026pvsr}}
& $<15^\circ$        & 16.76 & 25.15 & 24.16 & 37.47 \\
& $15^\circ$--$30^\circ$ & 21.16 & 26.29 & 30.23 & 37.09 \\
& $>30^\circ$        & 20.80 & 29.55 & 28.92 & 41.95 \\

\midrule

\multirow{3}{*}{PV-ASR~\cite{teng_2026pvsr}}
& $<15^\circ$        & 21.66 & 24.92 & 33.19 & 35.63 \\
& $15^\circ$--$30^\circ$ & 24.99 & 23.39 & 36.62 & 33.13 \\
& $>30^\circ$        & 26.49 & 28.18 & 38.27 & 40.22 \\

\midrule

\multirow{3}{*}{HP-VSR-Base (Ours)}
& $<15^\circ$        & 22.19 & 29.56 & 31.07 & 41.19 \\
& $15^\circ$--$30^\circ$ & 24.87 & 28.43 & 36.13 & 38.83 \\
& $>30^\circ$        & 24.73 & 31.33 & 36.61 & 43.23 \\

\midrule

\multirow{3}{*}{\makecell{HP-VSR-FiLMFuse \\ (L3--4) (Ours)}}
& $<15^\circ$        & 15.31 & 22.26 & 22.80 & 32.20 \\
& $15^\circ$--$30^\circ$ & 20.20 & 21.92 & 29.45 & 30.64 \\
& $>30^\circ$        & 19.82 & 24.97 & 28.14 & 35.52 \\

\midrule

\multirow{3}{*}{\makecell{HP-VSR-FiLMFuse \\ (L4) (Ours)}}
& $<15^\circ$        & 14.38 & 21.99 & 21.36 & 31.77 \\
& $15^\circ$--$30^\circ$ & 17.86 & 22.29 & 26.53 & 32.05 \\
& $>30^\circ$        & 17.78 & 24.52 & 25.12 & 35.59 \\

\midrule

\multirow{3}{*}{HP-VSR-ResFiLM (Ours)}
& $<15^\circ$        & 14.70 & 20.26 & 23.06 & 29.76 \\
& $15^\circ$--$30^\circ$ & 18.59 & 20.20 & 27.19 & 28.74 \\
& $>30^\circ$        & 17.84 & 22.74 & 26.29 & 33.10 \\

\bottomrule
\end{tabular}
}
\end{table*}

Methods without explicit pose-aware conditioning, such as V-ASR, exhibit noticeable performance degradation as head rotation increases. On LRS2, the WER of V-ASR rises from 24.16\% under near-frontal views to 28.92\% for poses exceeding $30^\circ$, while on LRS3 the WER increases from 37.47\% to 41.95\%. Although PV-ASR incorporates explicit landmark-based pose information and achieves improved overall performance, it remains sensitive to large pose variations. In particular, its LRS2 WER increases from 33.19\% to 38.27\%, while LRS3 WER rises from 35.63\% to 40.22\% as yaw exceeds $30^\circ$. These results suggest that landmark-based representations alone remain vulnerable to geometric distortion, self-occlusion, and viewpoint-dependent appearance changes introduced by large head rotations.

The proposed FiLM-based approaches demonstrate substantially improved robustness across all pose conditions. Compared with HP-VSR-Base, all FiLM variants achieve lower PER and WER across the three yaw ranges, indicating that pose-conditioned feature modulation effectively improves the quality of visual representations. Among the evaluated methods, HP-VSR-ResFiLM consistently achieves the strongest performance on LRS3 across all pose ranges. Under near-frontal conditions ($<15^\circ$), it achieves 29.76\% WER, compared with 32.20\% for HP-VSR-FiLMFuse (L3--4) and 31.77\% for HP-VSR-FiLMFuse (L4). The advantage persists under moderate poses ($15^\circ$--$30^\circ$), where HP-VSR-ResFiLM achieves 28.74\% WER compared with 30.64\% and 32.05\% for the two fusion-based variants, respectively.

As pose variation becomes more severe, the benefits of explicit pose-conditioned adaptation become increasingly apparent. Under large head rotations ($>30^\circ$), HP-VSR-ResFiLM achieves the lowest LRS3 WER of 33.10\%, outperforming both HP-VSR-FiLMFuse (L3--4) at 35.52\% and HP-VSR-FiLMFuse (L4) at 35.59\%. Similar trends are observed at the phoneme level, where HP-VSR-ResFiLM achieves the lowest LRS3 PER of 22.74\% under large-pose conditions. These results indicate that residual FiLM conditioning provides more effective pose-aware adaptation than direct deep-layer modulation, particularly when generalizing to challenging pose distributions.

On LRS2, the performance differences between FiLM variants are smaller. HP-VSR-FiLMFuse (L4) achieves the lowest WER across all yaw ranges, obtaining 21.36\%, 26.53\%, and 25.12\% WER for near-frontal, moderate, and large poses, respectively. However, the performance gap relative to HP-VSR-ResFiLM remains modest, suggesting that both approaches are effective for handling pose variation on LRS2. In contrast, the substantially larger improvements observed on LRS3 indicate that the benefits of residual FiLM conditioning become more pronounced under more challenging and diverse visual conditions.

These observations provide important insight into the role of FiLM-based conditioning. While explicit pose fusion combined with deep modulation can achieve strong performance on specific datasets and pose ranges, the proposed residual formulation exhibits greater consistency across pose conditions and datasets. By refining high-level visual representations through residual pose-aware adaptation while preserving the original feature manifold, HP-VSR-ResFiLM maintains stable performance from near-frontal to large-pose scenarios without requiring aggressive modulation of multiple deep layers.

Overall, the results demonstrate that explicitly incorporating head-pose information into the visual encoder substantially improves robustness to pose variation. Among the evaluated approaches, HP-VSR-ResFiLM provides the most consistent performance across datasets and pose conditions, while HP-VSR-FiLMFuse (L4) achieves the strongest results on LRS2. These findings further support the effectiveness of residual FiLM conditioning as a robust mechanism for pose-aware visual speech recognition in unconstrained real-world environments.

\section{Limitations and Future Work}\label{limits}

While the proposed framework demonstrates strong performance, several limitations highlight important directions for future research.

\begin{itemize}

\item \textbf{robustness under extreme pose variation.}
Although pose-conditioned FiLM modulation improves performance across a wide range of head poses, recognition accuracy still degrades under extreme yaw angles due to severe self-occlusion and asymmetric facial deformation. Ablation studies further reveal that the optimal modulation strategy depends strongly on pose severity. While the proposed residual FiLM formulation provides the best overall trade-off under standard viewing conditions, deeper Layer~3--4 modulation combined with explicit head-pose fusion becomes increasingly beneficial under large yaw angles exceeding $30^\circ$. This observation suggests that different pose regimes may require different levels of feature adaptation within the visual hierarchy. Future work will therefore investigate adaptive pose-aware modulation strategies that dynamically adjust the depth and strength of FiLM conditioning according to estimated pose magnitude. Potential directions include pose-adaptive gating mechanisms, hierarchical multi-stage FiLM conditioning, and geometry-aware representations that explicitly compensate for pose-induced distortions and self-occlusions.

\item \textbf{Intrinsic ambiguity of visual speech representations.}
Despite the use of phoneme-level modeling, visual speech recognition remains inherently ambiguous due to the many-to-one mapping between phonemes and visemes. This limitation indicates that improvements at the visual feature level alone may be insufficient to resolve ambiguity fully. Future research may benefit from integrating additional modalities or contextual cues and from designing more effective interfaces between visual recognition and language modeling to handle uncertainty in phoneme predictions better.

\item \textbf{Decoupled visual and linguistic modeling.}
The current two-stage framework isolates visual representation learning from language modeling to enable controlled analysis. However, this separation limits cross-stage interaction and may lead to suboptimal end-to-end performance. A promising direction is to explore tighter integration between visual and linguistic components, allowing joint optimization and improved error correction across stages. In particular, future work may investigate whether pose-aware visual conditioning can be jointly optimized with language-aware contextual refinement to improve robustness under challenging viewing conditions.

\item \textbf{Underexplored role of intermediate supervision signals.}
While CTC supervision is used during training, its potential for guiding inference and uncertainty estimation remains underutilized. This suggests a broader opportunity to leverage intermediate alignment signals for improving robustness, for example, through confidence-aware decoding, uncertainty-guided multimodal fusion, or hybrid inference strategies that dynamically combine multiple supervisory signals according to pose reliability and phonetic uncertainty.

\end{itemize}

\section{Conclusion}\label{conclusion}

This paper presented HP-VSR-ResFiLM, a pose-aware phoneme-based visual speech recognition framework that explicitly incorporates head-pose dynamics through FiLM-based feature modulation. By conditioning intermediate visual representations on estimated head pose, the proposed approach enables adaptive refinement of articulatory features under varying viewing conditions and improves robustness to pose-induced appearance changes. The framework adopts a two-stage architecture in which Stage~1 performs pose-aware phoneme prediction, while Stage~2 reconstructs word sequences using a fixed pretrained NLLB decoder. By keeping the linguistic component unchanged across experiments, the proposed framework enables controlled analysis of the contribution of pose-conditioned visual modeling. Extensive experiments on LRS2 and LRS3 demonstrate that the proposed approach consistently improves both phoneme-level and word-level recognition performance under comparable training conditions, outperforming prior phoneme-based approaches trained on the same datasets. The ablation studies further provide important mechanistic insights into the role of pose-conditioned modulation. In particular, the results show that the effectiveness of FiLM conditioning depends strongly on modulation placement and integration strategy. While deeper FiLM modulation combined with explicit head-pose fusion can provide strong performance on specific datasets, the proposed residual FiLM formulation achieves the strongest overall cross-dataset performance and the best balance between optimization stability, pose-aware adaptation, and recognition accuracy. Furthermore, the pose-based analysis demonstrates that residual FiLM conditioning maintains robust performance across a wide range of head orientations, indicating that adaptive pose-aware refinement effectively mitigates viewpoint-induced appearance variations and self-occlusion effects. Overall, the proposed framework demonstrates that explicitly incorporating head-pose information through residual FiLM-based feature modulation is an effective strategy for improving visual speech recognition under unconstrained pose variations. Beyond improving performance, the study also provides broader insight into how pose-aware conditioning interacts with hierarchical visual representations, offering a foundation for future research on adaptive, geometry-aware, and pose-robust visual speech recognition systems.

\section*{Declarations}

\subsection*{Ethical Approval}
This article does not contain any studies with human participants or animals performed by any of the authors.

\subsection*{Conflict of Interest}
The authors declare that they have no conflict of interest.

\subsection*{Funding}
This work was supported by JSPS KAKENHI Grant Number JP23H03787.

\subsection*{Data Availability}
The datasets used in this study include LRS2 \cite{son_2017lrs2} and LRS3 \cite{afouras_2018lrs3}, which are publicly available from their respective sources. 

\subsection*{Consent for Publication}
Not applicable.

\subsection*{Consent to Participate}
Not applicable.

\subsection*{Author Contributions}
Matthew Kit Khinn Teng: Conceptualization, Methodology, Software, Experiments, Writing -- Original Draft. \\
Haibo Zhang: Supervision, Writing -- Review \& Editing. \\
Takeshi Saitoh: Conceptualization, Supervision, Writing -- Review \& Editing. \\
All authors reviewed and approved the final manuscript.

\bibliography{sn-bibliography}

\end{document}